\pdfoutput=1

\documentclass[11pt]{article}

\usepackage[final]{acl}

\usepackage{times}
\usepackage{latexsym}

\usepackage[T1]{fontenc}

\usepackage[utf8]{inputenc}

\usepackage{microtype}

\usepackage{inconsolata}

\usepackage{graphicx}

\usepackage{amsmath}
\usepackage{multirow}

\usepackage{booktabs}

\usepackage{listings}
\lstdefinelanguage{yaml}{
  morestring=[b]',
  morestring=[b]",
  morecomment=[l]{\#},
  morekeywords={true, false, null},
  sensitive=true,
}

\usepackage{float} 

\usepackage{xcolor}
\usepackage{stfloats} 

\title{YaleNLP @ PerAnsSumm 2025: Multi-Perspective Integration via Mixture-of-Agents for Enhanced Healthcare QA Summarization}

\author{
  \textbf{Dongsuk Jang\textsuperscript{1,2,3}},
  \textbf{Alan Li\textsuperscript{1}},
  \textbf{Arman Cohan\textsuperscript{1}}\\
  \textsuperscript{1}Department of Computer Science, Yale University, \\
  \textsuperscript{2}Interdisciplinary Program for Bioengineering, Seoul National University, \\
  \textsuperscript{3}Integrated Major in Innovative Medical Science, Seoul National University\\
  \texttt{\{james.jang, haoxin.li, arman.cohan\}@yale.edu}  
}

\begin{document}
\maketitle
\begin{abstract}
Automated summarization of healthcare community question-answering forums is challenging due to diverse perspectives presented across multiple user responses to each question. The PerAnsSumm Shared Task was therefore proposed to tackle this challenge by identifying perspectives from different answers and then generating a comprehensive answer to the question. In this study, we address the PerAnsSumm Shared Task using two complementary paradigms: (i) a training-based approach through QLoRA fine-tuning of LLaMA-3.3-70B-Instruct, and (ii) agentic approaches including zero- and few-shot prompting with frontier LLMs (LLaMA-3.3-70B-Instruct and GPT-4o) and a Mixture-of-Agents (MoA) framework that leverages a diverse set of LLMs by combining outputs from multi-layer feedback aggregation. For perspective span identification/classification, GPT-4o zero-shot achieves an overall score of 0.57, substantially outperforming the 0.40 score of the LLaMA baseline. With a 2-layer MoA configuration, we were able to improve LLaMA performance up by 28\% to 0.51. For perspective-based summarization, GPT-4o zero-shot attains an overall score of 0.42 compared to 0.28 for the best LLaMA zero-shot, and our 2-layer MoA approach boosts LLaMA performance by 32\% to 0.37. Furthermore, in few-shot setting, our results show that the sentence-transformer embedding-based exemplar selection provides more gain than manually selected exemplars on LLaMA models, although the few-shot prompting is not always helpful for GPT-4o. The YaleNLP team’s approach ranked the overall second place in the shared task.
\end{abstract}

\begin{figure*}[!htb]
  \centering
\includegraphics[width=\linewidth]{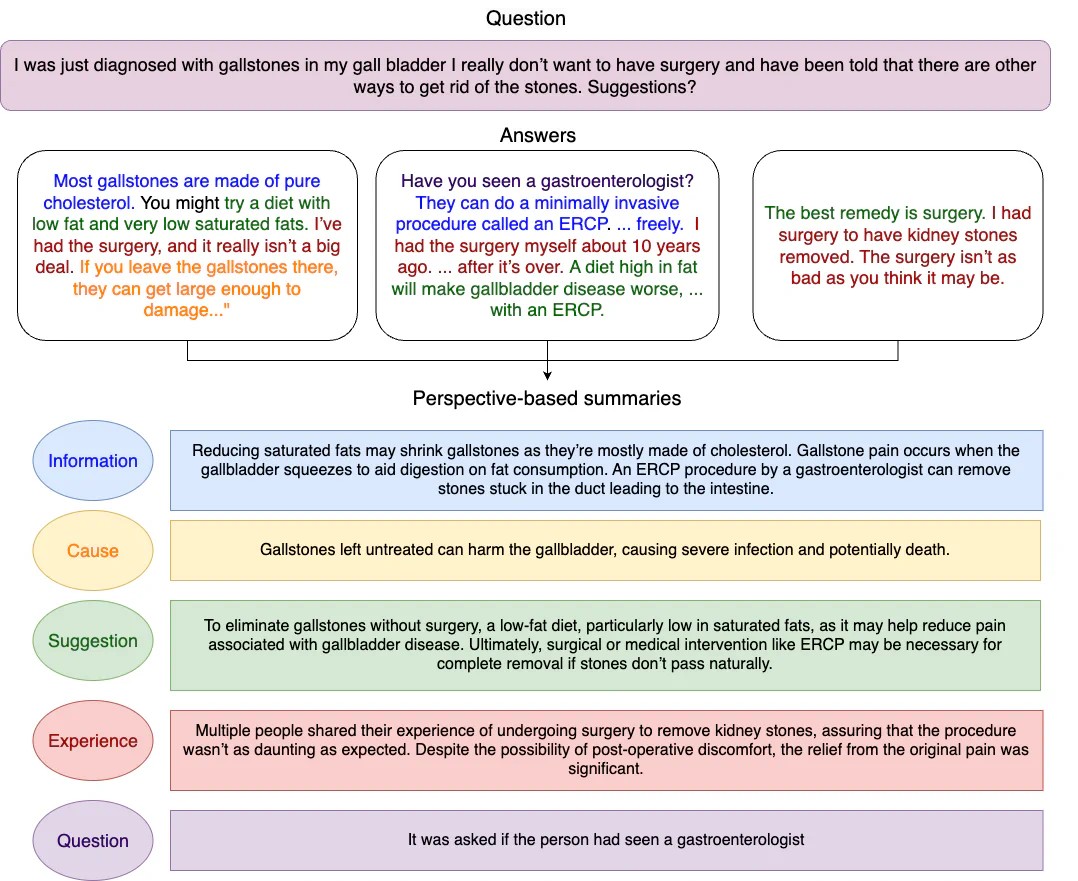}
  \caption{PerAnsSumm Shared Task overview.\protect\footnotemark}
  \label{fig:task}
\end{figure*}
\footnotetext{Figure adapted from \citeauthor{peransumm-overview} \url{https://peranssumm.github.io/docs/}}

\section{Introduction} \label{sec:intro}
Healthcare Community Question Answering (CQA) forums are rapidly growing as accessible platforms for individuals to seek medical advice, share personal experiences, or request simplified explanations of health conditions. Unlike expert-oriented medical sites, user-driven forums incorporate a broad range of viewpoints, from anecdotal evidence to speculative reasoning. Although the diversity can enrich the discussion, it also leads to information overload and frequent off-topic comments, making it difficult for newcomers to identify critical insights.
Traditionally, the CQA answer summarization task focuses on a single best-voted answer \cite{chowdhury2018cqasummbuildingreferencescommunity,chowdhury2020neuralabstractivesummarizationstructural} as a reference summary. However, a single answer often fails to capture the diverse perspectives presented across multiple answers. Providing the answers in structured, perspective-specific summaries could better serve the information needs of end users.

In response to this challenge, the \textbf{PerAnsSumm Shared Task} at the CL4Health@NAACL 2025 Workshop  \cite{peransumm-overview} introduces a perspective-specific summarization benchmark, encouraging researchers to design systems that explicitly recognize and integrate various user viewpoints into their outputs.
The task is comprised of two phases. Given a medical related query and a set of answers from CQA forums, the system is required to (i) identify the specific perspective in each of the answer and (ii) generate a summarization for each of these perspectives across different answers. Detailed task setup will be introduced in Section \ref{sec:task}.

Our main contributions and findings are as follows:
\begin{itemize}
    \item We show that GPT-4o \cite{openai2024gpt4ocard} generally outperforms 70B-level open-source models (the largest models we have access to) in both the span identification/classification and perspective-based summarization tasks. Providing few-shot examples do not \textit{consistently} yield higher performance.
    \item In few-shot setting, example selection by clustering on candidate example embeddings yield consistent improvements over manual example selection.
    \item Implementing a MoA \cite{wang2024mixtureofagentsenhanceslargelanguage} approach with multiple open-source LLMs significantly improves performance over individual models, demonstrating the potential of this ensemble strategy.
    \item QLoRA \cite{dettmers2023qloraefficientfinetuningquantized} fine-tuning with generic limited training data does not provide performance gains under our experimental conditions; in fact, it degrades performance. Due to the time constraints of the challenge, we were unable to explore additional fine-tuning configurations. We leave fine-tuning recipe exploration to future work.
\end{itemize}
We reimplemented\footnote{\url{https://github.com/JamesJang26/YALENLP-PerAnsSumm-2025}} the relevant techniques to align with the PerAnsSumm Shared Task. Through these experiments, our objective is to provide insight into the strengths and limitations of LLM-based approaches for perspective-aware summarization in medical CQA.

\section{Related Work}
Early abstractive summarization largely relied on pre-trained models such as BART \cite{lewis-etal-2020-bart}, T5 \cite{raffel2023exploringlimitstransferlearning}, and PEGASUS \cite{10.5555/3524938.3525989}, demonstrating strong performance on news benchmarks like CNN/DailyMail \cite{hermann2015teachingmachinesreadcomprehend} or XSum \cite{narayan-etal-2018-dont}. Yet, these approaches are typically optimized for well-structured, professionally written content. In contrast, healthcare forums contain personal opinions, anecdotal evidence, and multiple viewpoints that can hinder purely data-driven summarizers \cite{chaturvedi2024aspectorientedconsumerhealthanswer}.

Recent works in aspect- or perspective-oriented summarization highlight the value of parsing out different user viewpoints. \citet{naik-etal-2024-perspective} emphasize splitting content into categories like \textit{cause, suggestion, experience}, while AnswerSumm \cite{fabbri-etal-2022-answersumm} extracts sentence-level spans for query-focused summaries, though it does not fully address overlapping perspectives common in community Q\&A. At the same time, multi-document techniques for high-variance domains \cite{liu2018generatingwikipediasummarizinglong} suggest strategies for aggregating and reconciling disparate user responses.

Moreover, the rise of large language models has fueled interest in zero-/few-shot prompting, with studies showing that manually curated exemplars can be fragile or insufficiently general. Embedding-based selection methods like FsPONER \cite{tang2024fsponerfewshotpromptoptimization} and adaptive few-shot prompting \cite{tang2025adaptivefewshotpromptingmachine, Chang2021OnTI} propose retrieving exemplars via similarity or clustering, offering more stable and domain-sensitive prompts. Such techniques are well-suited to healthcare Q\&A, where a single misaligned exemplar can skew the summary toward incorrect or irrelevant details.

While prompt-based approaches can reduce reliance on large labeled datasets, certain tasks still benefit from specialized model tuning. To this end, LoRA \cite{hu2021loralowrankadaptationlarge} introduced a low-rank adaptation mechanism that updates only a small fraction of model parameters, and QLoRA \cite{dettmers2023qloraefficientfinetuningquantized} extends this concept by quantizing weights for further efficiency. These methods enable domain-focused tuning without the prohibitive resource costs typically associated with training massive LLMs from scratch.

\section{PerAnsSumm} \label{sec:task}
PerAnsSumm Shared Task is comprised of two subtasks sequentially, as shown in . Given a question $Q$, a set of answers $A$, and perspective categories \{\texttt{cause}, \texttt{suggestion}, \texttt{experience}, \texttt{information}, \texttt{question}\}, we are assigned the following two tasks:

\subsection{Task A: Span Identification and Classification}
For each answer in $A$, identify all text spans that convey any of the five perspectives. 

Following the task guidelines, systems must output a list of labeled spans. For example:
\begin{quote}
\texttt{span: ``<extracted span>", label: ``<perspective>"}
\end{quote}
Any text not relevant to a predefined perspective is omitted.

\paragraph{Evaluation Metrics}
\label{sec:eval_taskA}
PerAnsSumm evaluates Task~A under two main criteria:
\begin{itemize}
    \item \textbf{Classification}: Whether the model correctly assigns a perspective label to an answer if it contains that perspective. Macro-F1 and Weighted-F1 are reported.
    \item \textbf{Span Matching}: Compares predicted spans with gold-standard spans via strict matching and proportional matching. 
\end{itemize}
An overall macro-average of these measures is used for final ranking.

\subsection{Task B: Perspective-Based Summaries}
Building on Task~A, after identifying and labeling spans in a Q\&A thread, the system must produce a short, coherent summary for each perspective that appears.

Systems typically generate summaries in a structured format, for example:
\begin{quote}
\texttt{EXPERIENCE Summary: <text>}\\
\texttt{INFORMATION Summary: <text>}\\
\texttt{CAUSE Summary: <text>}\\
\texttt{SUGGESTION Summary: <text>}\\
\texttt{QUESTION Summary: <text>}
\end{quote}
If a perspective is deemed absent by the model, no summary is produced for that label.

\paragraph{Evaluation Metrics}
\label{sec:eval_taskB}
To assess each perspective-specific summary, PerAnsSumm combines measures of \emph{relevance} and \emph{factuality}, 
\begin{itemize}
\item \textbf{Relevance}: ROUGE \cite{lin-2004-rouge}, BLEU \cite{10.3115/1073083.1073135}, METEOR \cite{banerjee-lavie-2005-meteor}, and BERTScore \cite{zhang2020bertscoreevaluatingtextgeneration} quantify how well the generated summary aligns with the reference. 
\item \textbf{Factuality}: AlignScore \cite{zha-etal-2023-alignscore} and SummaC \cite{laban-etal-2022-summac} confirm that the summary is consistent with the original source text (i.e., it does not hallucinate or contradict).
\end{itemize}
These sub-metrics are aggregated into a final Task~B score.


\section{Methods} \label{sec:methods}
This section details the various modeling strategies we explore, including zero-/few-shot prompting, the MoA framework, and QLoRA supervised fine-tuning.

\subsection{Zero-/Few-Shot Prompting}
\label{subsec:prompting}

\paragraph{Zero-Shot Setup.}
We first experiment with prompting large language models using a instruction that specifies the task (either span identification/classification or perspective-based summarization). For instance, we provide definitions of the five perspectives (\textit{cause, suggestion, experience, information, question}) and ask the model to extract or summarize accordingly (best prompt for each tasks are detailed in Appendix~\ref{sec:prompt_example}). This approach requires no additional training or fine-tuning, leveraging the general knowledge embedded in instruction-tuned LLMs.

\paragraph{Few-Shot Setup.}
We provide 3--5 exemplars to the model via the prompt. We investigate two distinct methods for exemplar selection:
\begin{description}
    \item \textbf{Manually Curated}: We pick representative CQA threads that cover multiple perspectives and exhibit typical corner cases.
    \item \textbf{Embedding-Based Selection}: We embed all potential demonstration samples from training set with a sentence-transformer (e.g., \texttt{all-MiniLM-L6-v2} in our case), cluster them using $k$-means, and then pick top-$k$ samples based on proximity to the test query.
\end{description}

\subsection{Mixture-of-Agents}
\label{subsec:moa}
Medical content requires both domain knowledge and nuanced understanding of different viewpoints. To overcome these limitations, we implement a Mixture-of-Agents (MoA) framework that leverages the complementary strengths of multiple language models working in concert.
MoA is a framework for ensembling multiple sub-models (or agents) and integrating their outputs via an aggregator. We adapt and extend this method for our tasks. Specifically, we consider different numbers of layers (1, 2, or 3) in the MoA pipeline:
\begin{itemize}
\item \textbf{1-Layer MoA}: Each agent generates a partial response (e.g., predicted spans or short perspective-based summaries). An aggregator model then fuses these responses into a final output in a single step.
\item \textbf{2-Layer MoA}: After collecting agent outputs, we employ an intermediate "verification" layer to refine or check consistency before passing the refined results to the final aggregator model.
\item \textbf{3-Layer MoA}: We add an additional "hallucination detection" layer, which aims to filter out or correct unsupported statements before the final aggregation.
\end{itemize}
For our agent selection, we incorporate diverse models including open-source LLMs (LLaMA-3.3-70B-Instruct, Qwen-2.5-72B-Instruct, Deepseek-R1-Distill-LLaMA-70B) and closed-source models (GPT-4o, GPT-4o-mini). This diversity is intentional—each model brings different strengths in medical reasoning, language understanding, and factual recall. By combining them, we aim to create a system that outperforms any individual model, especially for complex medical content where perspectives might be subtle or require domain expertise.

We test various configurations to understand the optimal MoA architecture for each subtask. These configurations include combinations of open-source models only, GPT-4o only, and hybrid approaches where different model types handle different stages of the pipeline. For example, one effective arrangement uses GPT-4o for span identification/classification and a MoA ensemble for perspective-wise summarization based on those identified spans. We also explore the reverse configuration, as well as using MoA for both tasks. Through these experiments, we can measure the synergistic effects gained from mixing diverse LLMs and identify which models perform best at each stage of the process.

The multi-layer verification approach is particularly valuable for healthcare content, where accuracy is paramount. By adding verification and hallucination detection layers, we create checkpoints where potentially incorrect or unsupported information can be filtered or corrected before final aggregation, improving the reliability of the generated summaries.

\subsection{QLoRA Supervised Fine-Tuning}
\label{subsec:qlora}

While zero-/few-shot prompting relies on the generalization capabilities of LLMs, we also investigate QLoRA, a parameter-efficient fine-tuning approach. Through QLoRA, we can update a small set of low-rank adaptation parameters while keeping the majority of model weights frozen. This reduces both the computational overhead and memory usage compared to full fine-tuning.

\section{Experiments}
\label{sec:experiments}
We employ a diverse set of open-source models (LLaMA-3.3-70B-Instruct \cite{llama3-meta}, Qwen-2.5-72B-Instruct \cite{qwen2025qwen25technicalreport}, and Deepseek-
R1-Distill-LLaMA-70B \cite{deepseekai2025deepseekr1incentivizingreasoningcapability}) and closed frontier model (GPT-4o and GPT-4o-mini \cite{openai2024gpt4ocard}). 

\subsection{Experimental Data}
\label{subsec:dataset}
We employ the PUMA\footnote{\textbf{P}erspective s\textbf{UM}marization d\textbf{A}taset} \cite{naik-etal-2024-perspective} corpus provided by PerAnsSumm shared task. It contains 3,245 Q\&A threads, each with up to five perspective annotations (\texttt{cause, suggestion, experience, information, question}) and reference summaries per perspective. We follow official splits: 2,236 threads for training, 959 for validation, and 50 withheld for testing, while for the paper, we tested on the last 400 cases from valid set. 

\subsection{QLoRA Finetuning Implementation}
We used \texttt{llama-factory} \cite{zheng-etal-2024-llamafactory} toolkit to simply fine-tune LLaMA-3.3-70B-Instruct under various hyperparameter settings. For additional fine-tuning details, see Appendix~\ref{sec:qlora_appendix}.


\section{Results}
\label{sec:results}

\begin{table*}[!htb]
\centering
\small
\setlength{\tabcolsep}{4pt}
\renewcommand{\arraystretch}{1.2}
\begin{tabular}{llccccccccc}
\hline
\textbf{Model} & \textbf{Setting} & 
\textbf{M-F1} & \textbf{W-F1} & 
\textbf{St-P} & \textbf{St-R} & \textbf{St-F1} & 
\textbf{Pr-P} & \textbf{Pr-R} & \textbf{Pr-F1} & \textbf{Overall} \\
\hline
\multirow{4}{*}{LLaMA-3.3-70B-Instruct} 
  & Zero-shot           & 0.5381 & 0.7299 & 0.0320 & 0.1218 & 0.0507 & 0.4530 & 0.6991 & 0.5498 & 0.3968  \\
  & 3-shot w/ H    & 0.5390 & 0.7265 & 0.0339 & 0.1240 & 0.0513 & 0.4665 & 0.7163 & 0.5673 & 0.4031  \\
  & 3-shot w/ C & 0.5697 & 0.7676 & 0.0385 & 0.1311 & 0.0565 & 0.4954 & \underline{0.7404} & 0.5974 & 0.4246  \\
  & QLoRA SFT & 0.4788 & 0.6584 & 0.0256 & 0.1158 & 0.0447 & 0.4216 & 0.6681 & 0.5184 & 0.3664  \\
\hline
\multirow{3}{*}{GPT-4o} 
  & Zero-shot           & \textbf{0.8949} & \textbf{0.9190} & \textbf{0.1756} & \textbf{0.2641} & \textbf{0.2110} & \underline{0.6578} & 0.7392 & \underline{0.6961} & \textbf{0.5697}  \\
  & 3-shot w/ H    & 0.8176 & 0.8479 & \underline{0.1552} & 0.2193 & 0.1818 & 0.6145 & 0.7124 & 0.6599 & 0.5261  \\
  & 3-shot w/ C & \underline{0.8553} & \underline{0.8723} & 0.1468 & \underline{0.2546} & \underline{0.1862} & \textbf{0.6810} & \textbf{0.7525} & \textbf{0.7150} & \underline{0.5580}  \\
\hline
\multirow{2}{*}{MoA} 
  & Best 1             & 0.8129 & 0.8478 & 0.1491 & 0.2072 & 0.1734 & 0.5512 & 0.6942 & 0.6145 & 0.5063  \\
  & Best 2             & 0.7682 & 0.7809 & 0.1443 & 0.1697 & 0.1560 & 0.5412 & 0.6512 & 0.5912 & 0.4753  \\
\hline
\end{tabular}
\caption{\textbf{Task A}(span identification/classification) results . ``3-shot w/ H(uman)'' means three manually curated examples were used for few-shot prompting; ``3-shot w/ C(lustering)'' means three exemplars were automatically selected via sentence-transformer embeddings. Metrics include Macro-F1 (M-F1), Weighted-F1 (W-F1), Strict Precision/Recall/F1 (St-P, St-R, St-F1), Proportional Precision/Recall/F1 (Pr-P, Pr-R, Pr-F1), and an Overall average.}
\label{tab:taskA}
\end{table*}

\begin{table*}[!htb]
\centering
\small
\setlength{\tabcolsep}{4pt}
\renewcommand{\arraystretch}{1.2}
\begin{tabular}{llccccccccc}
\hline
\textbf{Model} & \textbf{Setting} & 
\textbf{R-1} & \textbf{R-2} & \textbf{R-L} & 
\textbf{BLEU} & \textbf{MET} & \textbf{BS} & 
\textbf{AS} & \textbf{SC} & \textbf{Overall} \\
\hline
\multirow{4}{*}{LLaMA-3.3-70B-Instruct} 
  & Zero-shot           & 0.2476 & 0.0886 & 0.2156 & 0.0471 & 0.2777 & 0.8182 & 0.3096 & 0.2247 & 0.2786  \\
  & 3-shot w/ H    & 0.2583 & 0.0968 & 0.2241 & 0.0487 & 0.2891 & 0.7612 & 0.2864 & 0.2345 & 0.2749  \\
  & 3-shot w/ C & 0.2733 & 0.0994 & 0.2398 & 0.0817 & 0.3055 & 0.8295 & 0.3151 & 0.2498 & 0.2993  \\
  & QLoRA SFT & 0.2165 & 0.0778 & 0.1947 & 0.0315 & 0.2460 & 0.7960 & 0.2486 & 0.2033 & 0.2518  \\
\hline
\multirow{3}{*}{GPT-4o} 
  & Zero-shot  & \textbf{0.4704} & \underline{0.2340} & \underline{0.4038} & \textbf{0.1307} & \textbf{0.4289} & \textbf{0.9116} & \textbf{0.4615} & \textbf{0.3031} & \textbf{0.4180}  \\
  & 3-shot w/ H    & \underline{0.4519} & 0.2291 & 0.3825 & 0.1193 & 0.3701 & 0.8821 & 0.4212 & 0.2543 & 0.3888  \\
  & 3-shot w/ C & 0.4515 & \textbf{0.2524} & \textbf{0.4057} & \underline{0.1212} & \underline{0.3987} & \underline{0.8901} & \underline{0.4552} & 0.2812 & \underline{0.4070}  \\
\hline
\multirow{2}{*}{MoA} 
  & Best 1             & 0.4372 & 0.2103 & 0.3611 & 0.1025 & 0.3305 & 0.8558 & 0.3913 & 0.2614 & 0.3688  \\
  & Best 2             & 0.4192 & 0.2055 & 0.3502 & 0.1096 & 0.3206 & 0.8512 & 0.3608 & \underline{0.2853} & 0.3628  \\
\hline
\end{tabular}
\caption{\textbf{Task B}(perspective-based summarization) results. ``3-shot w/ H(uman)'' vs.\ ``3-shot w/ C(lustering)'' follows the same few-shot definitions as Table~\ref{tab:taskA}. Metrics include ROUGE (R-1, R-2, R-L), BLEU, METEOR (MET), BERTScore (BS), AlignScore (AS), SummaC (SC), and an Overall average.}
\label{tab:taskB}
\end{table*}

We evaluate our approaches on two tasks: \textbf{~Task~A} (span identification/classification) and \textbf{~Task~B} (perspective-based summarization), using the macro-averaged metrics described in Section~\ref{sec:eval_taskA} and \ref{sec:eval_taskB}.

\subsection{Task A: Span Identification and Classification}

Table~\ref{tab:taskA} presents the classification and span-matching results.

\paragraph{GPT-4o Zero-Shot} remains the best overall single-model approach, scoring 0.5697 in Overall average, which notably outperforms all other models or methods. Detailed span identification results is described in Appendix~\ref{sec:gpt4o_confmat}.
\paragraph{Few-Shot Prompting} For both LLaMA-3.3-70B-Instruct and GPT-4o, embedding-based selection (0.4246 and 0.5580 overall) outperforms manually curated exemplars (0.4031 and 0.5261), showing better generalizability than human-chosen examples.

\paragraph{QLoRA Supervised Fine-tuning}
For Task A, our QLoRA-based fine-tuning of the LLaMA-3.3-70B-Instruct model (see Table~\ref{tab:taskA}) obtains an overall score of 0.3664, which is below the best zero- or few-shot baselines.

\begin{figure*}[!htb]
  \centering
  \includegraphics[width=\linewidth]{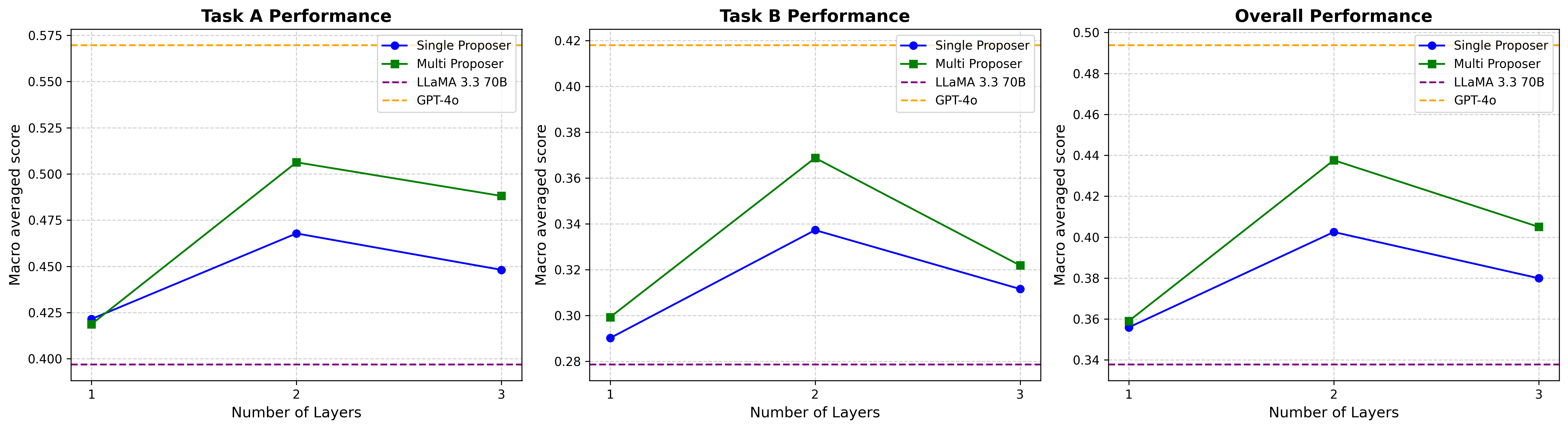}
  \caption{Performance comparison across different MoA layer counts.}
  \label{fig:moa_comparison}
\end{figure*}

\paragraph{MoA Details.}
\textbf{Best~1} is a 2-layer MoA with four open-source models in Layer~1 (two LLaMA-3.3-70B-Instruct + two Qwen-2.5-72B-Instrcut), one LLaMA-3.3-70B-Instruct in Layer~2, and an aggregator also based on LLaMA-3.3-70B-Instruct. 
As illustrated in Figure~\ref{fig:moa_comparison}, a 2-layer configuration strikes the best balance 
between thoroughness and retaining valid outputs, outperforming both 1-layer and 3-layer variants.
\textbf{Best~2} uses a similar 2-layer pipeline but swaps the sub-model composition to four temperature variants of LLaMA-3.3-70B-Instruct for Layer~1. Both surpass single LLaMA setups, underscoring MoA’s ability to fuse multiple perspectives effectively.

\paragraph{MoA} Best~1 at 0.5063, Best~2 at 0.4753 show strong improvements over single LLaMA-3.3-70B-Instruct baselines, though they still trail GPT-4o zero-shot. Nonetheless, MoA outperforms any single open-source LLM setting by a noticeable margin, 8\% over LLaMA's best.

\subsection{Task B: Perspective-Based Summaries}

Table~\ref{tab:taskB} shows the summarization performance, which is derived based on the best result spans from Task A, obtained using the optimal prompt detailed in Table~\ref{tab:prompt_structure} of Appendix~\ref{sec:prompt_example}. Once again, the zero-shot GPT-4o approach leads with a general average of 0.4180, exceeding its 3-shot variants and aligning with the trends observed in Task A.

\paragraph{Few-shot Prompting}
For LLaMA-3.3-70B-Instruct, ``3-shot w/ Clustering'' yields 0.2993 overall vs 0.2749 with human-chosen examples and 0.2786 in zero-shot. Similarly for GPT-4o, sentence-transformer embedding based selection attains 0.4070, surpassing the human-chosen 3-shot (0.3888) while still slightly lower than the zero-shot GPT-4o (0.4180). Hence, while GPT-4o with zero-shot remains the single best, sentence-transformer embedding based few-shot tends to outperform manually curated exemplars.


\paragraph{MoA}
\textit{Best~1} achieves 0.3688, while \textit{Best~2} gets 0.3628, each notably exceeding LLaMA-3.3-70B-Instruct’s best (0.2993). Although not rivaling GPT-4o, they confirm MoA’s capacity to reduce hallucinations and unify multiple sub-model outputs.

\paragraph{QLoRA Supervised Fine-tuning}
For Task B, QLoRA fine-tuning yields an overall score of 0.2518 (Table~\ref{tab:taskB}), again lower than the corresponding zero- and few-shot results. 

\subsection{Ablation on Aggregators and Layering}
\label{sec:moa_ablation}

\paragraph{Aggregator Comparison.}
Table~\ref{tab:aggregator_results} compares four aggregator models—LLaMA-3.3-70B-Instruct, Qwen-2.5-72B-Instruct, DeepSeek-R1-LLaMA-70B, GPT-4o-mini for the same MoA sub-model outputs (Best~1). LLaMA-3.3-70B-Instruct yields the highest Task~A/B scores (0.5063 / 0.3688), while the GPT-4o mini aggregator drops to (0.4027 / 0.2981), showing that the aggregator choice is crucial.

\begin{table}[!htbp]
\centering
\small
\setlength{\tabcolsep}{10pt}
\renewcommand{\arraystretch}{1.1}
\begin{tabular}{lcc}
\hline
\textbf{Aggregator} & \textbf{Task A} & \textbf{Task B} \\
\hline
LLaMA-3.3-70B-Instruct  & \textbf{0.5063} & \textbf{0.3688} \\
Qwen2.5-72B-Instruct    & \underline{0.4719} & \underline{0.3456} \\
DeepSeek-R1-LLaMA-70B   & 0.4671 & 0.3411 \\
GPT-4o-mini             & 0.4027 & 0.2981 \\
\hline
\end{tabular}
\caption{Performance comparison of different aggregators on Task A and Task B, holding the same MoA sub-model outputs as in ``Best\,1''.}
\label{tab:aggregator_results}
\end{table}

\paragraph{Layering Comparison.}
Figure~\ref{fig:moa_comparison} illustrates how adding layers impacts MoA performance under two configurations:
\begin{itemize}
    \item \textbf{Single Proposer:} Only LLaMA-3.3-70B-Instruct models are used to produce output in each layer.
    \item \textbf{Multi Proposer:} LLaMA-3.3-70B-Instruct and Qwen-2.5-72B-Instrcut are combined to generate more diverse proposals.
\end{itemize}
In both cases, LLaMA-3.3-70B-Instruct is used as the aggregator, and the dashed lines indicate the zero-shot baselines (LLaMA: 0.3377 overall; GPT-4o: 0.4938 overall).

In the \textbf{Single Proposer} setting, the 1-layer model obtains an overall score of 0.3559, which increases to 0.4025 with 2 layers (a gain of 0.0466 points) but then drops to 0.3799 when using 3 layers. Similarly, in the \textbf{Multi Proposer} setting, the overall score rises from 0.3590 for 1 layer to 0.4376 for 2 layers (an improvement of 0.0786 points), before falling to 0.4050 with 3 layers.

These results indicate that adding a second layer consistently improves performance—yielding an improvement of roughly 14\% over the LLaMA zero-shot baseline—while the third layer tends to over-correction, resulting in a performance drop. Thus, the 2-layer multi proposer configuration offers the best trade-off between enhancing overall accuracy and retaining valid outputs.

\section{Conclusion} In this work, we addressed the challenge of perspective-aware summarization for healthcare Q\&A. Our experiments show the recipe we tried and the final solution submitted for the challenge. With a bit disappointment, although MoA and embedding-based few-shot example selection improves the performance of open-source solution, the closed model, specifically GPT-4o in our case, still outperforms our best open-source solution by a large margin. Overall, our results highlight promising directions in leveraging large language models for multi-perspective healthcare Q\&A, particularly when curated resources are scarce. 

\section{Limitations} 
Data size and quality could be one of major constraints. The generic training set might be too small to conduct effective finetuning. 
In our observation, Text span identification/classification annotations contain overlaps and ambiguities (e.g. extracted span starts with an incomplete word or punctuation), complicating the accuracy of perspective labels and gold summaries. 

To apply an encoder-based model for span identification, we experimented with weighted NER fine-tuning (Appendix~\ref{sec:nerlike}). This approach assigns higher weights to underrepresented perspective categories to mitigate class imbalance. However, our results did not yield improvements, likely due to the inherent complexity and variability of user-generated content in the dataset. This suggests that alternative techniques, such as data augmentation or more robust fine-tuning strategies, may be necessary for handling imbalanced annotations effectively.

While MoA framework brings performance improvement, MoA configurations demand additional computational resources, especially in multi-layer or multi-agent setups. 

Addressing these limitations, for example, through larger, more balanced datasets and more efficient aggregator layers, could further enhance perspective-aware summarization in real-world healthcare scenarios.





\section{Future Work}

To overcome current constraints, future endeavors could involve extracting more healthcare-related queries from broader corpora such as Natural Questions, followed by data augmentation via LLMs to create synthetic examples for underrepresented perspectives. A refined Mixture-of-Agents design could then integrate these enriched training sets for both classification and summarization tasks, thereby mitigating data scarcity, enhancing perspective coverage, and improving model generalizability across diverse healthcare topics.

Although our preliminary exploration shows that embedding-based selection boosts performance over manually curated exemplars, further studies on prompting construction techniques, like dynamic prompt construction \cite{Gonen2022DemystifyingPI}, retrieval-augmented prompting \cite{tang2025adaptivefewshotpromptingmachine}, or synthetic prompts \cite{Kong2024QPOQP}, may lead to additional gains. We leave these investigations to future work, anticipating that such refinements will further enhance the robustness and scalability of perspective-aware summarization in the healthcare domain.

\section*{Acknowledgments}

This research was supported by a grant of the Korea Health Technology R\&D Project through the Korea Health Industry Development Institute (KHIDI), funded by the Ministry of Health \& Welfare, Republic of Korea (grant number : HI19C1352).

\bibliography{acl_latex}

\appendix

\section{Llama 3.3 70B QLoRA Supervised Fine-tuning Configs}
\label{sec:qlora_appendix}

As shown in Table~\ref{tab:qlora_config_table}, the following configuration was used for supervised fine-tuning using QLoRA for both Task A and Task B. The model was fine-tuned with 4 NVIDIA RTX A6000 GPUs. The only difference between the two tasks is the composition of the training dataset. This ensures that both tasks were fine-tuned under the same training environment, leveraging QLoRA to efficiently adapt the LLaMA-3.3-70B-Instruct model while maintaining computational efficiency.

\begin{table}[!ht]
\centering
\resizebox{\columnwidth}{!}{%
\begin{tabular}{ll}
\toprule
\textbf{Parameter} & \textbf{Value} \\
\midrule
bf16                    & true \\
cutoff\_len             & 3000 \\
dataset                & peranssumm\_task \\
dataset\_dir            & data \\
ddp\_timeout            & 180000000 \\
do\_train               & true \\
double\_quantization    & true \\
eval\_steps             & 5000 \\
eval\_strategy          & steps \\
finetuning\_type        & lora \\
flash\_attn             & auto \\
gradient\_accumulation\_steps & 2 \\
learning\_rate          & 5.0e-05 \\
logging\_steps          & 5 \\
lora\_alpha             & 16 \\
lora\_dropout           & 0.05 \\
lora\_rank              & 8 \\
lora\_target            & all \\
lr\_scheduler\_type     & cosine \\
max\_grad\_norm         & 1.0 \\
max\_samples            & 100000 \\
model\_name\_or\_path   & \{model\_name\} \\
num\_train\_epochs      & 3.0 \\
optim                   & adamw\_torch \\
output\_dir             & /path/to/output \\
packing                 & false \\
per\_device\_eval\_batch\_size & 1 \\
per\_device\_train\_batch\_size & 1 \\
plot\_loss              & true \\
preprocessing\_num\_workers & 16 \\
quantization\_bit       & 4 \\
quantization\_method    & bitsandbytes \\
report\_to              & none \\
save\_steps             & 5000 \\
stage                   & sft \\
template                & llama3 \\
train\_on\_prompt       & true \\
trust\_remote\_code     & true \\
val\_size               & 0.3 \\
warmup\_steps           & 100 \\
\bottomrule
\end{tabular}%
}
\caption{QLoRA Supervised Fine-Tuning Configuration for LLaMA-3.3-70B-Instruct}
\label{tab:qlora_config_table}
\end{table}

\section{Confusion Matrix for GPT-4o Zero-Shot}
\label{sec:gpt4o_confmat}

\begin{figure}[!h]
\centering
\includegraphics[width=1.0\linewidth]{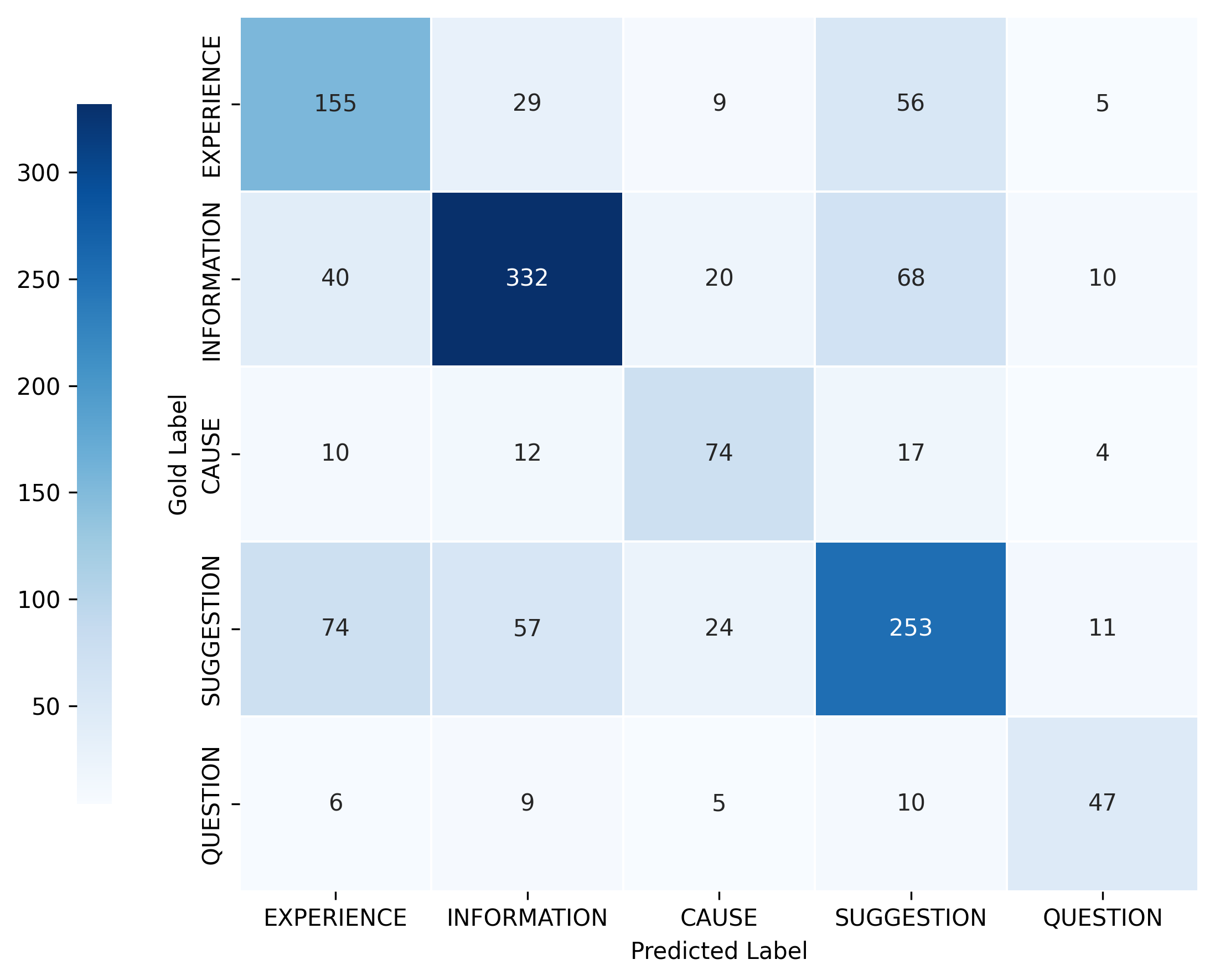}
\caption{Confusion Matrix for GPT-4o Zero-Shot on Task A. Each cell indicates the number of samples in the corresponding gold-predicted label pair.}
\label{fig:gpt4o_confmat}
\end{figure}

Figure~\ref{fig:gpt4o_confmat} shows the confusion matrix (as a PNG image) for GPT-4o zero-shot on Task A (span classification). Rows correspond to the \emph{gold} labels, and columns correspond to the \emph{predicted} labels. Diagonal entries represent correctly classified samples for each perspective category, whereas off-diagonal entries indicate misclassifications (e.g., gold-labeled \texttt{EXPERIENCE} predicted as \texttt{INFORMATION}).

As illustrated in the confusion matrix, GPT-4o zero-shot achieves strong diagonal counts for each perspective label (\texttt{EXPERIENCE}, \texttt{INFORMATION}, \texttt{CAUSE}, \texttt{SUGGESTION}, \texttt{QUESTION}), indicating accurate predictions in most cases. The off-diagonal cells reflect scenarios where one perspective is mistaken for another, highlighting specific patterns of confusion (e.g., \texttt{EXPERIENCE} vs.\ \texttt{INFORMATION}). This strong performance aligns with our earlier quantitative results showing that GPT-4o zero-shot outperforms other baselines on span classification.

\section{Prompt Example}

\label{sec:prompt_example}
In Table~\ref{tab:prompt_structure}, we present an example of the best prompt format for GPT-4o in zero-shot for both span identification/classification and perspective-based summarization.

\begin{table*}[t]
\centering
\small
\renewcommand{\arraystretch}{1.2} 
\setlength{\tabcolsep}{4pt} 
\begin{tabular}{|p{2.3cm}|p{5.8cm}|p{5.8cm}|}
\hline
\textbf{Task} & \textbf{Span Identification and Classification} & \textbf{Perspective-Based Summarization} \\
\hline
\textbf{System Prompt} & 
\multicolumn{2}{c|}{You are a helpful assistant.} \\
\hline
\textbf{User Prompt} & 
\begin{minipage}[t]{5.8cm}
\small
You are an expert annotator specialized in perspective-aware Healthcare Answer Summarization.  
\newline
First, validate that the document’s content is aligned with the medical domain—ensure that it pertains to prevention, diagnosis, management, treatment of diseases, understanding of bodily functions, the effects of medications or medical interventions, or queries regarding wellness practices.
\newline
Next, for each text span in the 'Answers' section, carefully assess and assign the most relevant perspective(s) from the following definitions:
\begin{itemize}
    \item \textbf{INFORMATION}: Knowledge about diseases, disorders, and health-related facts.
    \item \textbf{CAUSE}: Reasons responsible for the occurrence of a medical condition.
    \item \textbf{SUGGESTION}: Advice or recommendations to assist in making informed decisions.
    \item \textbf{EXPERIENCE}: Individual experiences or anecdotes related to healthcare.
    \item \textbf{QUESTION}: Inquiries for deeper understanding.
\end{itemize}
Follow these instructions:
\begin{itemize}
    \item Only annotate spans from the 'Answers' section.
    \item Ensure the document is medically relevant.
    \item Multi-perspective labeling is allowed.
    \item If a span explicitly mentions quantitative details, include that in your annotation.
    \item Avoid personal bias and exclude links or personal identifiers.
    \item Review your annotations to cover all underlying perspectives.
\end{itemize}
Format your response as: \texttt{span: "<extracted text>", label: "<perspective>"}.
\end{minipage} & 
\begin{minipage}[t]{5.8cm}
\small
While writing summaries, ensure that every essential idea and medical detail is captured from the extracted spans.  
\newline
Each summary should:
\begin{itemize}
    \item Be factually supported by the extracted spans.
    \item Preserve all relevant insights and details.
    \item Align clearly with the assigned perspective.
    \item Avoid hallucinations, bias, or unverifiable content.
\end{itemize}
Strictly adhere to the extracted spans to ensure factual consistency.
\newline
Use the following structure for each perspective:
\begin{itemize}
    \item \textbf{INFORMATION}: "For information purposes, [summary]..."
    \item \textbf{CAUSE}: "Some of the causes include [summary]..."
    \item \textbf{SUGGESTION}: "It is suggested that [summary]..."
    \item \textbf{EXPERIENCE}: "In user’s experience, [summary]..."
    \item \textbf{QUESTION}: "It is inquired whether [summary]..."
\end{itemize}
Format your final summary as: \texttt{Summary: "<generated summary>"}.
\end{minipage} \\
\hline
\textbf{Example Input} & 
\begin{minipage}[t]{5.8cm}
\small
\texttt{\{Question\} + \{Context\} + \{Answers\} + \{User Prompt\}}
\end{minipage} &
\begin{minipage}[t]{5.8cm}
\small
\texttt{\{Question\} + \{Context\} + \{Spans\} + \{User Prompt\}}
\end{minipage} \\
\hline
\end{tabular}
\caption{Final prompt structure for Task A and Task B.}
\label{tab:prompt_structure}
\end{table*}

\section{NER Fine-tuning for Task A}
\label{sec:nerlike}

In an exploratory experiment, we implemented a token-level BIO tagging\cite{ramshaw1995textchunkingusingtransformationbased} approach to perform span identification for Task A. In this method, each perspective is treated as a named entity with BIO labels (e.g., \texttt{B-INFORMATION}, \texttt{I-INFORMATION}, etc.), and the remaining tokens are tagged as \texttt{O}.

\paragraph{Data Preparation and Tagging.}
We first combined the question and answer texts and then tokenized the resulting sequence. Using the provided span annotations, we aligned token boundaries with the annotated spans to produce BIO tags. For instance, if an annotated span for the "CAUSE" perspective starts at character position \(s\) and ends at \(e\), tokens falling entirely within this span are labeled as \texttt{B-CAUSE} for the first token and \texttt{I-CAUSE} for the subsequent tokens.

\paragraph{Class Weighting for Imbalance.}
To address class imbalance, we computed class weights as:
\[
w_c = \frac{T}{n_c}, \quad \text{with } T = \sum_{c=1}^{C} n_c
\]
where \(n_c\) denotes the total number of tokens belonging to class \(c\), and \(T\) represents the total number of tokens across all classes. These weights were then incorporated into the cross-entropy loss function:  
\[
\mathcal{L} = - \frac{1}{N} \sum_{i=1}^{N} w_{y_i}\log \Biggl( \frac{\exp(z_{i,y_i})}{\sum_{c=1}^{C} \exp(z_{i,c})} \Biggr)
\]
where \(z_{i,c}\) is the logit for token \(i\) and class \(c\), and \(y_i\) is the ground-truth label.

\paragraph{Observations.}
Despite applying class weighting, our NER fine-tuning did not yield significant improvements. We attribute this to the small dataset size and the inherent challenge of labeling extended, overlapping spans—conditions that differ substantially from typical NER tasks involving shorter entity mentions. Consequently, while promising in principle, further investigation with larger or more targeted datasets is required.

\end{document}